# Evaluating Genetic Algorithms through the Approximability Hierarchy[1]

Alba Munoz~[a], Fernando Rubio[b,2]

[a]*Ingenium Research Group. ETSI Industriales*
*Universidad de Castilla-La Mancha, 13071 Ciudad Real, Spain* [b]*Instituto de Tecnología del Conocimiento*
*Dept. Sistemas Informaticos y Computación. Facultad de Informática*
*Universidad Complutense de Madrid, 28040 Madrid, Spain*

Abstract

Optimization problems frequently appear in any scientific domain. Most of the times, the corresponding decision problem turns out to be NP-hard, and in these cases genetic algorithms are often used to obtain approximated solutions. However, the difficulty to approximate different NP-hard problems can vary a lot. In this paper, we analyze the usefulness of using genetic algorithms depending on the approximation class the problem belongs to. In particular, we use the standard approximability hierarchy, showing that genetic algorithms are especially useful for the most pessimistic classes of the hierarchy.

*Keywords:* Heuristic methods, Genetic Algorithms, Complexity, Approximability.

1. Introduction

It is well known that many problems appearing in computational science are NPhard, meaning that we cannot provide algorithms finding their optimal solution in polynomial time unless P=NP. Let us recall that a problem belongs to *P* when it can be solved in polynomial time, while a decision problem belongs to *NP* when, given a candidate solution, we can check its correctness in polynomial time. Finally, a problem is NP-hard when it is at least as difficult to solve as any problem in *NP*.

[1] Work partially supported by projects TIN2015-67522-C3-3-R, PID2019-108528RB-C22, and by Comunidad de Madrid as part of the program S2018/TCS-4339 (BLOQUES-CM) co-funded by EIE Funds of the European Union.

[2] Corresponding author
*Email addresses:* albamu02@ucm.es (Alba Munoz),~ fernando@sip.ucm.es (Fernando Rubio)

Much research has been performed trying to provide alternative approximation algorithms for NP-hard problems. In particular, some good heuristics are specific to the problem under consideration (see e.g. [1, 2, 3, 4, 5, 6, 7]), whereas others are adaptations of general optimization metaheuristics. Among these metaheuristics, evolutionary computation methods [8] and swarm optimization methods [9], including genetic algorithms [10], particle swarm optimization [11], ant colony optimization [12], and many others (see e.g. [13, 14, 15, 16]), have proved their usefulness to deal with many optimization problems in the context of computational science (see e.g. [17, 18, 19, 20, 21, 22, 23, 24, 25, 26, 27, 28]). Even though these methods do not guarantee any target performance ratio in the worst case, they turn out to provide satisfactory results for typical problem instances.

However, although the difficulty of finding the optimal solution is the same for any NP-hard problem, the difficulty of finding *good* approximations can be very different for each problem. Given an approximation algorithm, we define its performance ratio as the ratio between the value of optimal solutions and the value of solutions found by the algorithm in the worst case. It has been proved that some NP-hard optimization problems can be approximated in polynomial time up to any performance ratio arbitrarily close to 1 (e.g. 0-1 Knapsack [29]), others can be approximated in polynomial time only up to some fix ratio (e.g. the best approximation ratio for MAX-3CNF is 7/8, unless P=NP [30]), and others cannot be approximated in polynomial time up to *any* constant performance ratio if $P \neq NP$ (e.g. Min Traveling salesman problem [31], Max Clique [32] or Min Set Cover [33]). In fact, during the last years it has been defined a detailed approximability hierarchy, and much effort has been done to classify the optimization versions of most of the classical NP-hard problems.

Taking into account the big differences regarding approximability between different optimization problems, we wonder if the usefulness of using metaheuristics (compared to using ad-hoc methods) depends on the approximation class a problem belongs to. In order to deal with a concrete generic heuristic, we will compare the usefulness of using genetic algorithms comparing it with the results obtained using problem-specific heuristics, trying to answer questions like: Is it worth to use genetic algorithms in case the problem belongs to a *good* approximation class? And in case it belongs to an *intermediate* or *bad* approximation class? In what situations should we use it?

In order to answer the previous questions, we will analyze a representative of each approximation class. For each problem, we will use a standard benchmark to compare the results obtained by a well-known standard ad-hoc heuristic for such problem, and the results obtained using a genetic algorithm. Moreover, we will also consider the case where the results of the ad-hoc heuristic can be used to improve the quality of the genetic algorithm. Let us remark that we are not interested in finding the optimal configuration to be used to deal with each problem when using a genetic algorithm, because our main aim is to analyze the basic trends. Thus, we will use a simple genetic algorithm in every case.

Our experimental results suggest that the usefulness of genetic algorithms increases as the approximability of the problem worsens. In fact, for the most optimistic approximation class the genetic algorithm will not provide valuable improvements over ad-hoc heuristics, while in the most pessimistic approximation class a simple genetic algorithm will obtain results that are much better than those obtained by the ad-hoc heuristics used.

Let us remark that many authors have already analyzed the performance of genetic algorithms by taking into account different aspects. For instance, the performance of different genetic operators and the interaction between them have been extensively studied in the literature (see e.g. [34, 35, 36]), including also the application in the context of continuous domains (see e.g. [37]). Moreover, empirical studies have also considered the usefulness of these methods in different application domains (see e.g. [38, 39, 40, 41]). However, to the best of our knowledge, a systematic empirical study about its usefulness depending on the approximability hierarchy has never been done before.

Some authors have already combined genetic algorithms and the approximability hierarchy, but only in the context of justifying the difficulty of dealing with a given problem. Good examples are [42, 43, 44, 45, 46, 20, 47], where the authors first prove the approximability class a problem belongs to, and then they provide a genetic algorithm to solve it. However, there has been no previous attempt to analyze the effectiveness of genetic algorithms depending on the approximation class. Thus, it seems that the research questions we deal with in this paper have not been analyzed before: For what approximation classes is it worth to use genetic algorithms alone? In what classes is it better to combine genetic algorithms with ad-hoc methods? In what cases is it better to consider only ad-hoc methods? We will provide empirical results to help answering these questions.

2. Brief introduction to the approximability hierarchy

As stated in the introduction, polynomial approximation is used to achieve feasible solutions whose objective value is as close as possible to the optimal value in polynomial time. In this section, polynomial approximation is going to be used to classify the optimization versions of NP-hard problems into different classes according to the best performance ratios that can be found for them (unless P=NP).

In order to measure the quality of the approximation algorithms there are mainly two paradigms:

- Standard approximation: the performance of the approximation algorithm A is evaluated by the ratio:

$$\rho_A(I) = \frac{v_A(I,S)}{opt(I)} \qquad (1)$$

where $v_A(I, S)$ is the objective value of the solution $S$ for the instance $I$ calculated by the algorithm A and $opt(I)$ is the optimal value for the instance $I$. This performance ratio is in $[1, \infty)$ for minimization problems and in $(0, 1]$ for maximization problems.

- Differential approximation: the performance of the approximation algorithm A is evaluated by the ratio:

$$\delta_A(I) = \frac{|\omega(I) - v_A(I,S)|}{|\omega(I) - opt(I)|} \quad (2)$$

where $\omega(I)$ is the worst solution of the instance *I*. This performance ratio is in [0, 1] for minimization and maximization problems.

Let us note that depending on the paradigm used, the classification (based on the performance ratios) is going to result into different classes (see e.g. [48]). For the sake of clarity, in this paper we will only deal with the standard approximation.

Therefore, based on their best possible performance ratios (resulting from the standard approximation), NP-hard problems are classified into different *approximability* classes. Named from the worst ratios to the best, the *approximability* classes are:

- Exp-APX: NP-hard problems whose best possible ratio is exponential (or the inverse of an exponential in maximization problems[1]) to the size of the instance. The well known *Minimum TSP* for complete graphs belongs to this class.

- Poly-APX: to this class belong NP-hard problems whose best possible ratio is polynomial (or the inverse of a polynomial in maximization problems) to the size of the instance. The *Maximum Independent Set* and *Minimum Coloring* problems are in Poly-APX.

- Log-APX: to this class belong NP-hard problems whose best possible ratio is logarithmic (or the inverse of a logarithmic in maximization problems) to the size of the instance. The *Minimum Set Covering* and *Minimum Dominating Set* problems are in Log-APX.

- APX: with this class begin the most optimistic classes. Here belong NP-hard problems whose best possible ratio is a fixed constant. The *Minimum Vertex Cover* and *Minimum Metric TSP*[2] are members of APX.

- PTAS: to this class belong NP-hard problems with a *polynomial-time approximation scheme*. A polynomial-time approximation scheme is an algorithm with an approximation ratio of 1 + $\varepsilon$ (1 - $\varepsilon$ for maximization problems) in polynomial time to the size of the instance for every fixed $\varepsilon$ > 0. The *Minimum Euclidean TSP* [3], *Minimum Planar Vertex Cover* and *Maximum Planar Independent Set* problems belong to PTAS.

- FPTAS: to this class belong NP-hard problems with a *fully polynomial-time approximation scheme*, this is a polynomial-time approximation scheme that is also polynomial with $1/\varepsilon$. The *Knapsack* problem is the best known problem in this class.

---

[1] The ratio in maximization problems is in (0, 1].
[2] Minimum TSP in complete graphs where the intercity distances satisfy the triangle inequality.
[3] Minimum TSP in complete graphs where the intercity distance is the euclidean distance.

These classes are related as follows:

FPTAS ⊂ PTAS ⊂ APX ⊂ Log-APX ⊂ Poly-APX ⊂ Exp-APX

where inclusions are strict unless P=NP.

3. Experimental results

For each approximation class, in this section we describe a canonical representative, we implement an ad-hoc heuristic solving it, and we also provide a genetic algorithm solving the same problem. In all cases, the genetic algorithm has been executed 20 times, and we show the best result obtained, the average result, and the standard deviation. For each problem, we provide results for three instances of increasing size that have been obtained from standard benchmarks. For each instance, we compare the results obtained by the ad-hoc heuristic, the genetic algorithm (GA), and by the GA when the output of the ad-hoc heuristic is provided as an initial individual of the GA. In all cases, we use the same number of individuals (100) and the same number of generations (500). Thus, we always limit the number of function evaluations to 50,000, so that the comparisons between methods are fair. Moreover, statistical tests are used to study whether the results obtained with different methods are statistically significant or not.

The pseudocode of the basic genetic algorithm used for the experiments can be seen in Algorithm 1, where the termination condition depends on reaching a fixed number of generations. For each problem under consideration, a different fitness function has to be used, while the crossover and mutation functions can also be different to take into account the concrete representation of the individuals in each problem. The only common function for every problem is the selection function, for which tournament selection has been chosen (its implementation is described in Algorithm 2).

---

Algorithm 1 Pseudocode for genetic algorithm

1: $t \leftarrow 0$
2: *initializePop(P(t)); Initializes population*
3: while not termination do
4:     *eval(P(t)); Evaluates population*
5:     P'(t) ← select(P(t)); Select best individuals in population
6:     P''(t) ← crossover(P'(t)); Generate offspring from best individuals
7:     P(t+1) ← mutate(P''(t)); Mutate some individuals, next generation is finished
8:     $t \leftarrow t + 1$
9: end while

---

Algorithm 2 Pseudocode for tournament selection method
---
1: for $j \leftarrow 0$ to $N$ do
2:     $\{i_1,...,i_k\} \leftarrow$ Choose k individuals from the population at random
3:     $i'_j \leftarrow$ Select the best evaluated individual in the tournament as the winner
4: end for
5: return $\{i'_1, ..., i'_N\}$
---

### 3.1. FPTAS: Knapsack problem

We start analyzing the most optimistic class, FPTAS. We have chosen the knapsack problem as representative of the class, as it is a very well-known and studied problem.

It is defined as follows.

**Definition 1 (Knapsack problem).** *Given a set of n items, each with a value ($v_i \geq 0$) and a weight ($w_i > 0$) associated, and a knapsack with a maximum weight $W > 0$, determine the subset of items that maximizes the value of the knapsack and whose total weight is less than or equal to W. In other words, find the subset $\{x_1, x_2,..., x_n\}$ that:*

$$\text{maximize} \sum_{i=1}^{n} v_i x_i \qquad \text{subject to} : \sum_{i=1}^{n} w_i x_i \leq W \qquad (3)$$

*where $x_i = 1$ if the item i is included in the knapsack and $x_i = 0$ otherwise.*

A very simple greedy algorithm for the knapsack problem is to put in the knapsack the *best* item left until there is no room for any of the remaining objects. Thus, the best item to put into the knapsack is going to be the one that maximizes the value and minimizes the weight, or maximizes the ratio between value and weight. We are going to call this ratio density:

$$density(i) = \frac{v_i}{w_i} \qquad (4)$$

So, the steps of the greedy algorithm are (i) sorting objects by density; (ii) select the object with the highest density and check if it fits in the knapsack; (iii) if the object fits in the knapsack, add it to the solution, if not, discard the object; and (iv) if there are any objects left, go back to step (ii).

Regarding the genetic algorithm, we use a basic implementation using tournament selection, with 100 individuals running during 500 iterations. Table 1 summarizes the results obtained for a set of benchmarks obtained from [49]. Three instances of increasing size (from 100 to 500 items) have been tested. For each of them, we compare the obtained results with that of the optimal solution for the corresponding instance. In the case of the genetic algorithm, we show the average result obtained after 20 independent executions, the standard deviation and the best result obtained. As it can be seen, the genetic algorithm (GA) can find the optimal solution (distance 0% to the optimal solution) in the simplest instance, outperforming the ad-hoc heuristic. However, GA's behavior worsens in the case of the largest instances. In fact, the ad-hoc heuristic obtains much better results in that case.

The last three columns of Table 1 show the results obtained when one of the initial individuals of GA is obtained by using the ad-hoc heuristic. By doing so, the solution of the corresponding GA will always be at least as good as that of the ad-hoc heuristic, and it could be better in case the GA can improve this solution in subsequent iterations. As it can be seen in the table, this combination GA + ad-hoc obtains the best results in any case, but the difference with respect to using only the ad-hoc method is quite small (even zero in the hardest instance), being the ad-hoc greedy method much faster. In fact, Figure 1 shows the evolution in time of the (average) quality of the solutions found by GA alone for each of the instances, using the execution time of the ad-hoc method as the basic time-unit. That is, we show the quality of the solution of GA when using the same amount of time required by the ad-hoc method, 250 times the execution time of the ad-hoc method, 500 times, and so on. As it can be seen, even in the easiest instance, GA needs 3,000 times the execution time of the ad-hoc method to outperform it. Moreover, in the hardest instance, even after 30,000 times longer executions, the ad-hoc method still outperforms the results of GA.

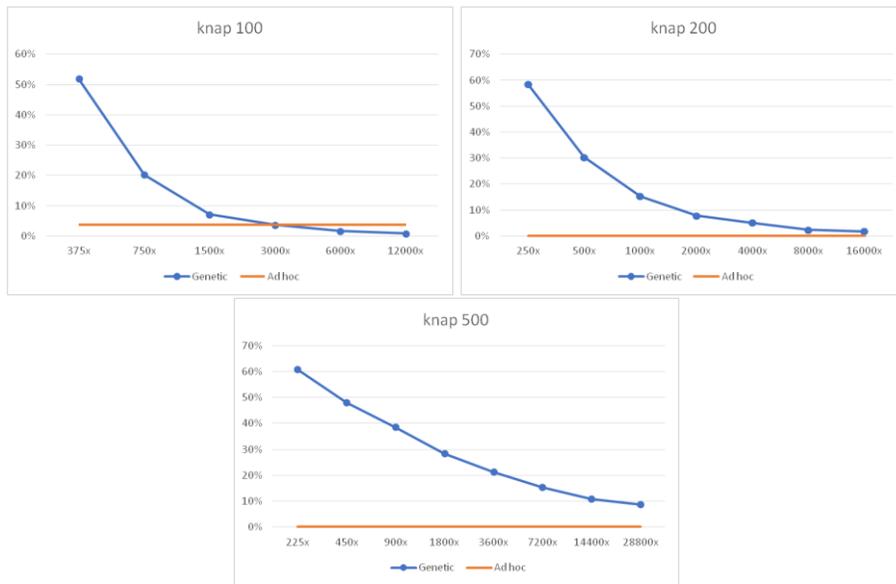

Figure 1: GA vs ad-hoc: evolution on time for knapsack problem

In order to be sure about the advantage of the genetic + ad-hoc method compared to the genetic algorithm alone, we have computed a non-paremetric test. In particular, for each of the three instances, we have used an independent Mann-Whitney U test to compare the 20 results obtained by the genetic algorithm alone, with the results obtained in the 20 executions of the genetic + ad-hoc method. The test concludes that in the simplest instance both methods behaves analogously (the p-value obtained is 0.25), while in the other two instances the significance tests indicate that the genetic + ad-hoc method behaves clearly better than the genetic algorithm alone (p-values are in both cases less than 0.0001).

Table 1: FPTAS-Knapsack: Overcost with respect to the optimal solution

| Instance | Ad-hoc | Genetic | | | Genetic + Ad hoc | | |
|---|---|---|---|---|---|---|---|
| | | Mean | Std | Best | Mean | Std | Best |
| knap 100 | 3.6% | 1.6% | 1.56 | 0.0% | 1.17% | 1.25 | 0.0% |
| knap 200 | 0.09% | 5.26% | 2.63 | 0.48% | 0.08% | 0.03 | 0.0% |
| knap 500 | 0.08% | 28.32% | 3.55 | 20.92% | 0.08% | 0.0 | 0.08% |

*3.2. PTAS: Euclidean Traveling Salesman Problem*

Next we deal with the representative of the PTAS class. In this case, we consider the euclidean version of the traveling salesman problem (TSP).

Definition 2 (Euclidean TSP). *Given a graph G = {V, E} where V = {$c_1$, $c_2$,..., $c_n$} is the set of vertices (each vertex corresponds to one city) and E is the set of edges, let d: E −→ R be the distance function that assigns to each edge ($c_i$, $c_j$) the euclidean distance between the cities $c_i$ and $c_j$. Find the shortest possible route that visits each city and returns to the first city visited.*

In this case we will use a two-steps ad-hoc algorithm. First, we apply a greedy algorithm to obtain a first solution. Then, we use the 2-OPT algorithm [50] to improve the quality of the solution. The initial greedy algorithm gradually builds the path by choosing in each step the shortest edge that can be added fulfilling the following conditions:

- When adding the edge, do not form cycles of less than *n* edges, being *n* the total number of nodes of the graph.

- When adding the edge, the degree of the vertices already added to the path is stillless than or equal to two.

Thus, the greedy algorithm works as follows: (i) sort edges by distance; (ii) select the shortest edge and add it to the route if it meets the above requirements; (iii) if the number of edges in the route is less than *n* go to step (ii).

The 2-OPT algorithm starts with a Hamiltonian cycle (obtained with the previous greedy algorithm) and goes through the cycle comparing the distance between the current edges $(v_i, v_j)$, $(v'_i, v'_j)$ with the distance that would result from crossing them, that is, the distance between $(v_i, v'_j)$, $(v'_i, v_j)$. If the new distance is less than the previous one, it makes the corresponding exchange of edges.

Regarding the genetic algorithm, we use again a basic implementation using tournament selection, with 100 individuals running during 500 iterations. Table 2 summarizes the results obtained for a set of benchmarks obtained from TSPLIB [51].

As in the case of knapsack, we show the results obtained for three instances of increasing size (ranging from 29 to 76 cities). As it can be seen in the table, GA can outperform 2-OPT in the case of the simplest instance, while the results obtained for the largest instances are worse. However, the combination of both methods clearly outperforms both of them. That is, by using the output of 2-OPT as an initial individual, it helps the GA focusing its search, obtaining much better results with the same number of function evaluations of the GA. In order to be sure about this advantage, we follow the same strategy described in the case of the knapsack problem. That is, we use a Mann-Whitney U test for each of the instances. The results of the significance tests indicate that there is not a statistical difference between both methods in the case of the simplest instance (the p-value is 0.13), but there is a clear statistical difference in the rest of the cases (p-values are less than 0.0001).

Thus, we conclude that the best option in this case is to use GA providing as input the solution obtained by 2-OPT.

*3.3. APX: Minimum Vertex Cover*

The Minimum Vertex Cover (MVC) is a typical representative of the APX class. The definition is as follows.

Table 2: PTAS-Euclidean TSP: Overcost with respect to the optimal solution

| Instance | Ad-hoc | Genetic | | | Genetic + Ad hoc | | |
|---|---|---|---|---|---|---|---|
| | | Mean | Std | Best | Mean | Std | Best |
| sahara | 25.5% | 29.79% | 15.66 | 1.59% | 23.19% | 14.57 | 4.7% |
| berlin52 | 13.4% | 43.72% | 10.43 | 24.21% | 8.43% | 0.85 | 7.25% |
| pr76 | 35.3% | 107.99% | 15.25 | 71.21% | 25.49% | 3.51 | 18.27% |

Definition 3 (MVC). *Given a graph G = {V, E} a vertex cover of G is a subset of vertices in V that includes at least one vertex of each edge in E. Therefore, given a graph, the Minimum Vertex Cover problem consists in finding the vertex cover of that graph with the smallest cardinal.*

In this case, being E the set of edges of the graph, the ad-hoc heuristic works as follows: (i) Initialize the cover C as an empty set; (ii) sort the vertices of the current graph from highest to lowest degree; (iii) add the highest-grade vertex *u* to C; (iv) remove from E all edges that contain *u*; and (v) if E is not empty, go back to (ii).

Regarding the genetic algorithm, we use again a basic implementation using tournament selection, with 100 individuals running during 500 iterations. Table 3 summarizes the results obtained for a set of benchmarks obtained from BHOSLIB [52]. As in the previous cases, we show the results obtained for three instances of increasing size (from 450 to 760 vertices). In this case the GA outperforms the ad-hoc heuristic in any case. Moreover, providing the output of the ad-hoc heuristic as an individual of the GA can still be an interesting idea: Although the best result obtained can be found by the GA alone, the average results obtained are better when using the result of the ad-hoc heuristic as an individual of the GA. The advantage in this case seems to be smaller than in the case of Euclidean TSP. However, when using a Mann-Whitney U test to compare the results obtained using both strategies we find out that the significance test confirms that in all instances there is a clear statistical difference between both methods (in all cases p-values are less than 0.001).

All in all, the results obtained by GA alone are worse than those obtained by GA with the cooperation of the ad-hoc heuristic, although in some cases the best result can be found by the GA alone.

Table 3: APX-Minimum Vertex Cover: Overcost with respect to the optimal solution

| Instance | Ad-hoc | Genetic | | | Genetic + Ad hoc | | |
|---|---|---|---|---|---|---|---|
| | | Mean | Std | Best | Mean | Std | Best |
| frb30-15-1 | 2.14% | 1.67% | 0.38 | 0.95% | 1.32% | 0.28 | 0.71% |
| frb35-17-1 | 1.78% | 1.7% | 0.25 | 0.89% | 1.09% | 0.16 | 0.71% |
| frb40-19-1 | 2.36% | 1.51% | 0.26 | 0.83% | 1.19% | 0.16 | 0.97% |

*3.4. Log-APX: Minimum Set Covering*

The best-known problem belonging to Log-APX is the Minimum Set Covering (MSC) problem, that is defined as follows.

**Definition 4 (MSC).** *Given a set X and a family S = {$S_1$, $S_2$,..., $S_m$} of subsets of X, a set covering of X with S is a set I ⊆ {1, 2,..., m} of indices such that $\bigcup_{j \in I} S_j$ = X. Thus, the Minimum Set Covering problem is, given a set X and a family S of subsets of X, to determine the set covering of X with S whose cardinal is the smallest.*

The ad-hoc heuristic adds at each step the subset that covers more elements of X. Thus, the steps are the following: (i) Initialize the cover C as an empty set; (ii) compute the subset T of S that covers the largest number of elements that have not been covered yet; (iii) add T to C; (iv) if there are still uncovered items, go to step (ii).

Regarding the genetic algorithm, we use again a basic implementation using tournament selection, with 100 individuals running during 500 iterations. Table 4 summarizes the results obtained for a set of benchmarks obtained from the OR-Library [53, 54]. As in the previous cases, we show the results obtained for three instances of increasing size (from 511 elements and 210 subsets, to 2047 elements and 495 subsets). In all cases, the differences between GA and the ad-hoc method are really large. Thus, using GA in this case is clearly a good idea. Regarding the combination of GA and the ad-hoc method, the results are equal to those obtained in the case of MVC. That is, the best solution found by the genetic algorithm alone can be better than the best solution found in cooperation with the ad-hoc heuristic, but the average result obtained is worse. Moreover, when we use the Mann-Whitney U test we find out that for all instances the significance test indicates that GA alone behaves clearly worse than in cooperation with the ad-hoc heuristic.

Table 4: Log-APX-Minimum Set Covering: Overcost with respect to the optimal solution

| Instance | Ad-hoc | Genetic | | | Genetic + Ad hoc | | |
|---|---|---|---|---|---|---|---|
| | | Mean | Std | Best | Mean | Std | Best |
| clr10 | 128% | 36.4% | 7.44 | 24% | 28% | 4.1 | 24% |
| clr11 | 152% | 57.39% | 7.8 | 34.78% | 48.91% | 3.11 | 39.13% |
| clr12 | 246% | 47.5% | 9.84 | 30.76% | 39.42% | 5.96 | 23.07% |

*3.5. Poly-APX: Maximum Independent Set*

In this case, we deal with the Maximum Independent Set (MIS) problem, whose definition is given below.

Definition 5. *Given a graph G = {V, E} an independent set is a subset of vertices in V where no two of them are adjacent. In other words, for every two vertices in the independent set there is no edge in E connecting them. Then, given a graph G, the Maximum Independent Set problem is to find the independent set of G with the largest cardinal.*

In the *Maximum Independent Set* problem, the aim is to add as many vertices as possible to the independent set to make it maximum. That is why in each step the adhoc algorithm will add to the set the vertex whose degree is minimum: (i) Initialize the auxiliary set $W$ with all the vertices in $V$ and the independent set $S$ as empty; (ii) select the vertex $v \in W$ of lesser degree. (iii) add $v$ to $S$ and remove from $W$ both $v$ and all its adjacents; and (iv) if there are any vertices left in W, go back to step (ii).

Regarding the genetic algorithm, we use again a basic implementation using tournament selection, with 100 individuals running during 500 iterations. Table 5 summarizes the results obtained for a set of benchmarks obtained from BHOSLIB [52]. As in the previous cases, we show the results obtained for three instances of increasing size (from 450 to 760 vertices). Again, the differences between GA and the ad-hoc method are really large in all cases. However, the combination of GA and the ad-hoc method does not seem to have a clear advantage over the GA alone. In fact, for each of the instances we have used again a Mann-Whitney U test, and it concludes that there is not a statistical relevant difference between both methods in any of the instances. That is, both methods behave equally well. Thus, using the ad-hoc method as input of the GA does not help the GA at all.

Table 5: Poly-APX-Maximum Independent Set: Overcost with respect to the optimal solution

| Instance | Ad-hoc | Genetic | | | Genetic + Ad hoc | | |
|---|---|---|---|---|---|---|---|
| | | Mean | Std | Best | Mean | Std | Best |
| frb30-15-1 | 33.3% | 22% | 3.96 | 10% | 20.66% | 3.99 | 13.33% |
| frb35-17-1 | 34.3% | 21.99% | 4.36 | 14.28% | 22.42% | 3.74 | 11.42% |
| frb40-19-1 | 33.3% | 22.37% | 2.36 | 15% | 22.37% | 3.76 | 10% |

*3.6. Exp-APX: Traveling Salesman Problem*

In this case we have the same definition as in the case of the Euclidean TSP, the only difference being that now the length of each edge is not restricted to be the euclidean distance between both nodes. Let us note that with that *simple* distinction, the two versions of the problem belong to very different approximability classes: euclidean TSP belongs to PTAS (a very optimistic class) while non-metric TSP belongs to ExpAPX (one of the classes with the worst approximation ratio).

The ad-hoc heuristic used to solve the non-metric TSP is the same as the one used in Section 3.2 to deal with the euclidean TSP. The GA is also the same, using 100 individuals during 500 iterations. Again, we compare the results obtained for three graphs of increasing size (from 20 to 50 nodes). However, in this case we do not know the optimal solution. Thus, instead of comparing the overcost with respect to the optimal solution, we directly compare the quality of the solutions found by each method, as shown in Table 6. The situation is similar to the one obtained in the Poly-APX case. That is, GA clearly outperforms the ad-hoc strategy. Moreover, the collaboration between the ad-hoc heuristic and the GA does not improve the final results. In fact, using again the Mann-Whitney U test for each of the instances we obtain that there is not a statistical relevant difference between both methods in any instance. As both methods behave equally well, it is not interesting to use the ad-hoc method as an input of the GA.

*3.7. Discussion*

If we analyze as a whole the complete set of tables presented in this section, we can see a clear trend. When a problem belongs to a class that has bad approximability, the usefulness of genetic algorithms is much higher than when it belongs to an optimistic class. Let us remind that the approximability classes are related as follows (assuming $P \neq NP$):

$$FPTAS \subset PTAS \subset APX \subset Log\text{-}APX \subset Poly\text{-}APX \subset Exp\text{-}APX$$

Table 6: Exp-APX-TSP: Best solutions found

| Instance | Ad-hoc | Genetic | | | Genetic + Ad hoc | | |
|---|---|---|---|---|---|---|---|
| | | Mean | Std | Best | Mean | Std | Best |
| matrix 20 | 388 | 275.05 | 40.67 | 163 | 248.7 | 38.77 | 159 |
| matrix 35 | 739 | 390.7 | 53.87 | 248 | 420.4 | 53.11 | 282 |
| matrix 50 | 1428 | 680.4 | 59.41 | 492 | 655.05 | 48.16 | 523 |

Our experiments suggest that using ad-hoc strategies can be a better option when we deal with problems belonging to FPTAS. In these cases, ad-hoc strategies can be much faster than genetic algorithms, while the solutions found by genetic algorithms are not significantly better.

When we deal with problems in PTAS, genetic algorithms start to be more useful. However, they can need a little bit of help. In fact, in these situations the best option is to provide genetic algorithms where some initial individuals are computed by using ad-hoc heuristics. By doing so, ad-hoc heuristics help the genetic algorithm to focus on interesting areas of the search space.

In the case of classes APX and Log-APX, the best option is the same as in PTAS. That is, providing individuals computed by using ad-hoc heuristics help improving the quality of the results obtained by using the GA. However, in these cases we only obtain advantages in the average case, while the best solution can sometimes be found by the GA working alone.

Finally, genetic algorithms are clearly useful in Poly-APX and Exp-APX. In these classes, genetic algorithms obtain much better results than the corresponding ad-hoc heuristic. Moreover, the usefulness of using ad-hoc heuristics to create initial individuals for the genetic algorithm is not relevant. The reason is that using this strategy can sometimes worsen the results, as they can easily conduct the genetic algorithm to focus the search very prematurely around areas representing local minima. That is, we can use this strategy, but we do not obtain better results on average. Thus, it is recommended to use the GA alone, as it obtains the same results with less effort.

4. Conclusions

The approximability hierarchy classifies optimization problems depending on the difficulty that we have to guarantee good approximations when we solve them. In this paper, we have analyzed its influence on the usefulness of genetic algorithms. Our experiments suggest that genetic algorithms are very useful to deal with any optimization problem provided that it belongs to any class ranging from PTAS to Exp-APX. However, when dealing with problems belonging to PTAS, it is recommended to provide implementations where the genetic algorithm collaborates with other ad-hoc heuristics.

In fact, this cooperation is also recommended for APX and Log-APX classes (although in these cases genetic algorithms can also obtain good results working alone after several trials). Finally, in the case of problems belonging to FPTAS, it is not so useful to use genetic algorithms, as similar results can be obtained faster using ad-hoc heuristics. These results are summarized in Table 7.

Table 7: Summary of results: Winner for each approximation class

| Approximation class | Example | Ad-hoc | Genetic + ad-hoc | Genetic |
|---|---|---|---|---|
| FPTAS | Knapsack | X | | |
| PTAS | Euclidean TSP | | X | |
| APX | Minimum vertex cover | | X | |
| Log-APX | Minimum set covering | | X | |
| Poly-APX | Maximum independent set | | | X |
| Exp-APX | Traveling Salesman Problem | | | X |

We conclude that whenever we confront a new optimization problem, we should start studying the approximation class it belongs to. That is, it is not enough to know that the corresponding decision problem is NP-hard: Having information about its approximability will allow us to select the appropriate way of using ad-hoc heuristics, genetic algorithms or combinations of both methods.

Ackwnowledments

The authors would like to thank the anonymous reviewers for valuable suggestions on a previous version of this paper.

Biosketches

Alba Munõz is a researcher at the Ingenium Research Group in Universidad de Castilla La Mancha (Spain). She obtained her BS degree in Mathematics in 2020 and she is currently finishing a MS Degree on Artificial Intelligence. Her research interests cover swarm and evolutionary optimization methods, formal methods, and complexity theory.

Fernando Rubio is an Associate Professor in the Computer Systems and Computation Department, Complutense University of Madrid (Spain). He obtained his MS degree in Computer Science in 1997, and he was awarded by the Spanish Ministry of Education with "Primer Premio Nacional Fin de Carrera". He finished his PhD in the same subject four years later. Dr. Rubio received the Best Thesis Award of his faculty in 2001. Dr. Rubio has published more than 90 papers in international refereed conferences and journals. His research interests cover formal methods, swarm and evolutionary optimization methods, parallel computing, and functional programming.